\pdfoutput=1

\documentclass[11pt]{article}

\usepackage{acl}

\usepackage{times}
\usepackage{latexsym}
\usepackage{inconsolata}
\usepackage{amsmath}
\usepackage{amsfonts}
\usepackage{amssymb}
\usepackage{graphicx}
\usepackage{multirow}
\usepackage{booktabs}
\usepackage{diagbox}
\usepackage{bbding}
\usepackage{algpseudocode}
\usepackage{enumitem}
\usepackage{commath}
\usepackage{algorithm}
\usepackage[toc,page]{appendix}
\usepackage{tabularx}
\usepackage{listings}
\usepackage{colortbl}
\usepackage{xcolor}
\usepackage{comment}
\usepackage[T1]{fontenc}

\usepackage[utf8]{inputenc}
\usepackage{csquotes}

\usepackage{microtype}

\usepackage{inconsolata}

\title{Cascading Large Language Models for Salient Event Graph Generation}

\author{Xingwei Tan$^{1,2}$, Yuxiang Zhou$^2$, Gabriele Pergola$^1$, Yulan He$^{1,2,3}$ \\
  $^1$Department of Computer Science, University of Warwick, UK\\
  $^2$Department of Informatics, King's College London, UK\\
  $^3$The Alan Turing Institute, UK\\
  \texttt{\{Xingwei.Tan, Gabriele.Pergola.1\}@warwick.ac.uk}\\
  \texttt{\{Yuxiang.Zhou, Yulan.He\}@kcl.ac.uk}\\}

\begin{document}

\maketitle
\begin{abstract}
Generating event graphs from long documents is challenging due to the inherent complexity of multiple tasks involved such as detecting events, identifying their relationships, and reconciling unstructured input with structured graphs.
Recent studies typically consider all events with equal importance, failing to distinguish salient events crucial for understanding narratives.
This paper presents CALLMSAE, a CAscading Large Language Model framework for SAlient Event graph generation, which leverages the capabilities of LLMs and eliminates the need for costly human annotations.
We first identify salient events by prompting LLMs to generate summaries, from which salient events are identified. 
Next, we develop an iterative code refinement prompting strategy to generate event relation graphs, removing hallucinated relations and recovering missing edges. 
Powered by CALLMSAE, we present \textit{NYT-SEG}, a large-scale automatically annotated event graph dataset which can serve as distant supervision signals.
Fine-tuning contextualised graph generation models on \textit{NYT-SEG} outperforms the models trained on CAEVO data.
Results on a human-annotated test set show that the proposed method generates salient and more accurate graphs, outperforming competitive baselines. 
\footnote{Code and data: \url{https://github.com/Xingwei-Tan/CALLMSAE}.}

\end{abstract}

\section{Introduction}

\label{sec:intro}

Events are fundamental discourse units which form the backbone of human communication. 
They are 
interconnected through various event relations such as hierarchical, temporal, or causal relations. 
Event relation graphs are vital for representing and understanding complex event narratives, with nodes representing events and edges denoting relationships between them.
High-quality event relation graphs can enhance numerous downstream tasks, such as question answering \cite{lu-etal-2022-event} and reasoning \cite{melnyk-etal-2022-knowledge}.

\begin{figure}
    \centering
    \includegraphics[width=0.95\columnwidth]{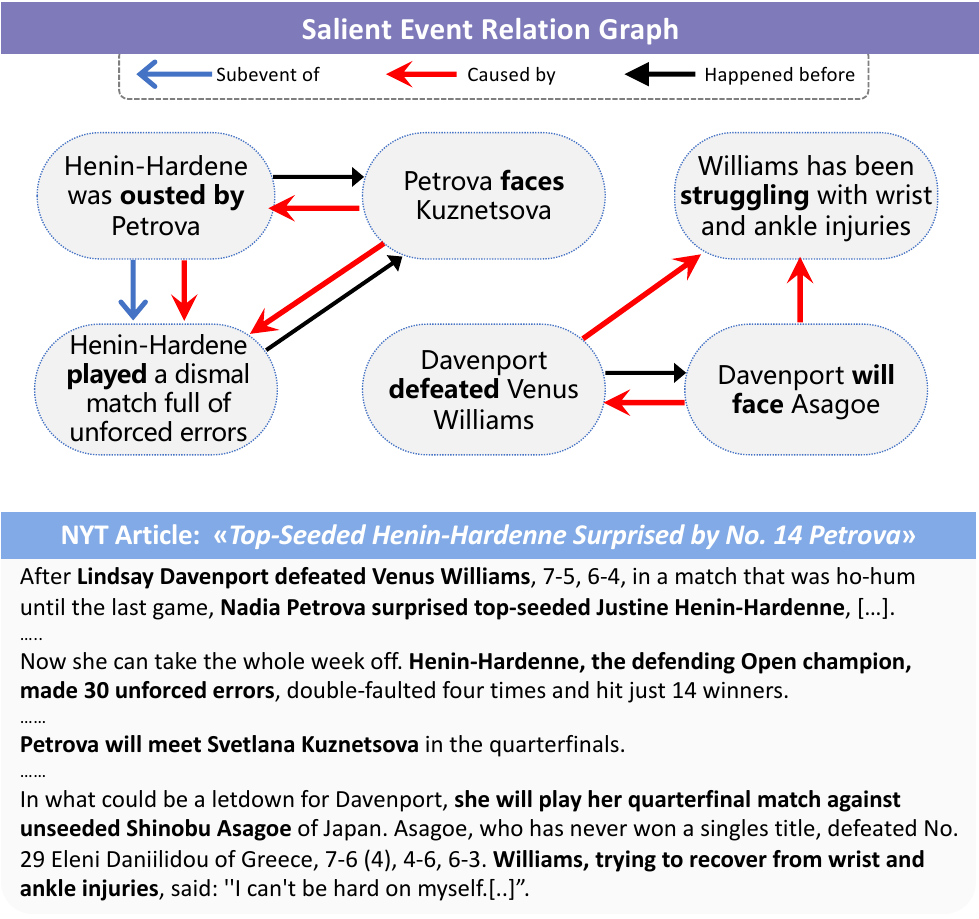}
    \caption{An example of salient event relation graph (top) generated from the NYT article (bottom). \vspace{-12pt}}
    \label{fig:example_inputoutput}
\end{figure}

Existing studies on contextualized event graph generation primarily focus on fine-tuning language models for end-to-end generation of linearized graphs from documents \cite{madaan-yang-2021-neural,tan-etal-2024-set}.
These methods rely on distant supervision, such as events and event temporal relations detected using CAEVO \cite{mcdowell-etal-2017-event}, due to the data-intensive nature of language models and the significant manual effort required for annotating event graphs.  
However, CAEVO's bottom-up, predicate-centric extraction often results in sparse, low-quality event graphs populated with non-salient events. For example, CAEVO frequently identifies all verbs as events, including trivial ones like "\emph{say}" and "\emph{think}," which have minimal connections to other events and contribute little to narrative understanding, thus introducing noise.  
In recent research involving human annotation \cite{dunietz-gillick-2014-new,liu-etal-2018-automatic}, a \textit{summarisation test} is leveraged to identify salient events or entities, where an event or entity is considered salient if a human-written summary is likely to include it.   

Motivated by these insights, we aim to improve the quality of distant supervision and mitigate noise in event graph generation by incorporating event saliency. 
Our hypothesis is that effective graph generation benefits from a top-down strategy, where the main events are identified first, rather than extracted solely in a bottom-up manner based on the predicate.
Capitalizing on the powerful summarization capabilities of large language models (LLMs), we propose a method that first instructs LLMs to summarize documents before identifying salient events.
Specifically, we introduce a code-prompting strategy for constructing event relation graphs encompassing hierarchical, temporal, and causal relations\footnote{We extend beyond the CAEVO's temporal-only relations to encompass multiple relation types.} (see Figure \ref{fig:example_inputoutput}).
Unlike vanilla prompting, which queries each potential event pair individually, our code prompt format generates each relation type in a single pass.  Furthermore, we incorporate an iterative refinement process using a \textit{hallucination grader} to filter spurious edges and iterative generation to recover missing ones. Finally, the abstractive nature of salient events presents an evaluation challenge, as they rarely match gold standard events exactly, even when semantically equivalent. To address this, we propose evaluation metrics based on semantic text embeddings for assessing the generated event relation graphs.

Moreover, we extend beyond the CAEVO's temporal-only relations to encompass multiple relation types.
We introduce iterative refinement prompting in a code prompt format to generate event relation graphs that include hierarchical, temporal, and causal relations (see Figure \ref{fig:example_inputoutput}).
The prompting framework is highly efficient because the code prompt format generates each type of relation graph in a single pass,  while the naive prompting method needs to query each possible event pair individually.
The iterative refinement process further enhances the accuracy of event relation predictions by using a hallucination grader to filter out unfaithful edges and iterative generation to recover missing edges.

Using the LLM-generated dataset, we fine-tune Flan-T5 following the same method as \citet{tan-etal-2024-set}.
However, the abstractive nature of salient events poses challenges for evaluation, as salient events rarely exactly match the gold standards despite having the same semantic meaning.
To address this, we propose an evaluation metric based on semantic text embeddings for assessing the event relation graphs.
Our experimental results on the New York Times corpus \cite{AB2/GZC6PL_2008} show that CALLMSAE, a novel CAscading Large Language Model framework for SAlient Event graph generation, outperforms the baselines in terms of event saliency and edge quality.
The fine-tuned model surpasses previous models trained with CAEVO-generated graphs.
Our contributions are summarised as follows:
\begin{itemize}
    \item We propose CALLMSAE, a CAscading Large Language Model framework for SAlient Event graph generation, serving as a distant signal generator for contextualised graph generation models.
    \item We propose a novel contextualised evaluation metric for comparing salient event graphs. Our extensive experimental evaluation on LLM-generated event relation graphs in terms of event saliency and event relation on the NYT corpus, demonstrating how higher quality salient event graphs can improve contextualised graph generation.
    \item We provide a large-scale LLM-generated salient event graph dataset \textit{NYT-SEG} with three major event relation types for distant supervision ($10,231$ documents), along with a human-annotated test set ($100$ documents).  
\end{itemize}

\section{Related Work}
\paragraph{Event Relation Graph Construction}
The early idea of event relation graph construction comes from \citet{uzzaman-etal-2013-semeval}, which introduces a dataset for evaluating an end-to-end system which takes raw text as input and output TimeML annotations (i.e., temporal relations).
CAEVO \cite{mcdowell-etal-2017-event} and Cogcomptime \cite{ning-etal-2018-cogcomptime} both utilise a wide range of manually designed features to train MaxEnt and averaged perception for extracting events and relations.
\citet{han-etal-2019-joint} proposed a joint event and relation extraction model based on BERT \cite{devlin-etal-2019-bert} and BiLSTM \cite{panchendrarajan-amaresan-2018-bidirectional}.
Other researchers focus on developing specialised sub-systems to classify extracted event pairs for relations \cite{ning-etal-2019-improved,han-etal-2019-deep,wang-etal-2020-joint,pergola-etal-2021-disentangled,pergola-etal-2021-boosting,tan-etal-2021-extracting, tan-etal-2023-event}.
ATOMIC \cite{10.1609/aaai.v33i01.33013027} is a large-scale commonsense knowledge graph containing the causes and effects of events.
MAVEN-ERE \cite{wang-etal-2022-maven} is built with event coreference, temporal, causal and subevent relations.
However, ATOMIC and MAVEN-ERE completely rely on crowdsourcing and thus are difficult to extend.
MAVEN-ERE is less than half the size of our dataset and does not consider the saliency of events.

\citet{madaan-yang-2021-neural} fine-tune GPT-2 to generate linearised graphs from documents in an end-to-end manner.
Their temporal relation graphs used for training are produced by CAEVO.
Following this direction, \citet{tan-etal-2024-set} instead view the task as set generation and propose a framework based on set property regularisation and data augmentation.
In this paper, we focus on generating multi-relation graphs via in-context learning, prompt interaction, and iterative refinement.

\paragraph{Salient Event Identification}
Several existing papers investigate the problem of identifying salient events.
\citet{choubey-etal-2018-identifying} build a rule-based classifier to identify central events by exploiting human-annotated event coreference relations.
They find the central events either have large numbers of coreferential event mentions or have large stretch sizes.
\citet{jindal-etal-2020-killed} propose a contextual model to identify salient events based on BERT and BiLSTM.
They also mention several features, such as event trigger frequency, which are essential features to identify the salient events.
\citet{liu-etal-2018-automatic} propose a feature-based method using LeToR \cite{liu2009learning} and a neural-based method called Kernel-based Centrality Estimation.
To train and evaluate their methods, they build a dataset based on the \textit{summarisation test}: an event is considered salient if a summary written by a human is likely to include it.
\citet{zhang-etal-2021-salience} combine the Kernel-based Centrality Estimation with the event and temporal relation extraction model of \citet{han-etal-2019-joint} to build a salience-aware event chain modelling system.
However, they only focus on single-dimensional chains and only model temporal relations.

\begin{figure*}
    \centering
    \includegraphics[width=0.99\linewidth]{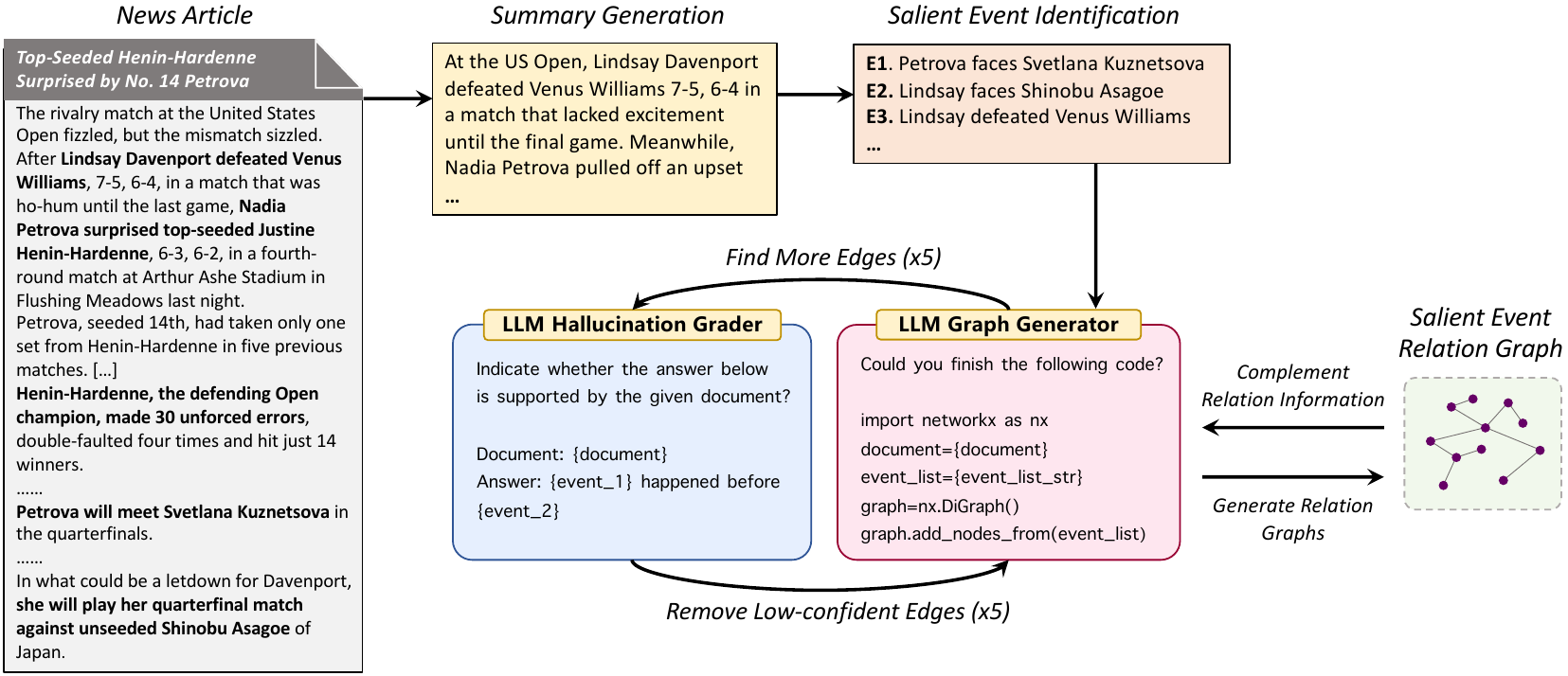}
    \caption{The proposed CALLMSAE framework.}
    \label{fig:main_method}
\end{figure*}

\section{Cascading LLMs to Generate Salient Event Graphs}

CALLMSAE combines various prompts in a pipelined manner to generate salient event graphs.
In this section, we introduce the prompts for generating salient events, and then the method for generating relation graphs based on the salient events.
Lastly, we define an evaluation metric for comparing event graphs: \emph{Hungarian Graph Similarity}.

\subsection{Generate Salient Events}
\label{sec:salient_event_extraction}

The \textit{summarisation test} (Section \ref{sec:intro}) is often used to guide the annotation of salient events or entities \cite{dunietz-gillick-2014-new,liu-etal-2018-automatic,lu-etal-2023-napss,lyu24}.
These studies identify events or entities included in human-written summaries as salient.
Similarly, we instruct LLMs to generate a summary first and then extract events from it. 
Examples of the summarisation and salient event generation prompts are shown in the Appendix (Table \ref{summary_prompt_table} and \ref{event_prompt_table}).

\subsection{Generate Graphs as Code Completion}

While LLMs can extract salient events, they often struggle with identifying event relations  \cite{chan2023chatgpt,sun-etal-2022-phee,sun-etal-2024-leveraging,tan-etal-2024-set}.
Prompt engineering for extracting event relations is complex due to the need to incorporate various terminologies and graph constraints.
Moreover, prompt efficiency is crucial as generating a large-scale dataset with LLMs can still incur significant computational costs, albeit less than crowdsourcing.

In our method, the main prompt for generating the event relation graph is formulated as a Python code completion task.
The graph is defined using the NetworkX\footnote{\url{https://networkx.org/documentation/stable/}} package in Python, with nodes representing the salient events generated in Section \ref{sec:salient_event_extraction}.
LLMs are instructed to complete the code by adding relation edges using NetworkX's APIs.

Recent research suggests that formulating prompts as code can enhance LLMs' reasoning abilities \cite{wang-etal-2023-code4struct,zhang-etal-2023-causal}. 
In our task, the Python code format effectively incorporates all necessary terminologies, enabling LLMs to understand them without confusion.
The Python code format also allows for the inclusion of constraints (e.g., ensuring the graph is a directed acyclic graph) and additional instructions (e.g., ask for explanations) as comments.
LLMs can generate explanations as comments without disrupting the main content of the graph, which is difficult to achieve in other formats (e.g., JSON).
Moreover, the code template simplifies parsing the response, as LLMs are directed to use the ``.add\textunderscore edge()'' function to add the relations.

Since hierarchical, temporal, and causal relations are asymmetric, each can be represented by a Directed Acyclic Graph (DAG).
We formulate three distinct prompts to guide LLMs in generating three DAGs, each representing one of these relation types. 
This approach avoids the complexity of a multi-label graph, and 
LLMs can focus on a single relation type and carefully consider the topological structure of the graph. 
We can also use the ``.find\textunderscore cycle()'' function from NetworkX to detect constraint violations reliably.
In addition, if relation types are interconnected, the initially generated graphs can help the generation of subsequent graphs (as will be explained in Section \ref{progressive_gen}).
We provide an example of the code prompt in Appendix (Table \ref{prompt_table}).

\subsection{Iterative Refinement}
\label{sec:grader}
\paragraph{Hallucination Grader} The code prompt efficiently guides LLMs to generate graphs, but it often generates hallucinated relations.
Based on our preliminary experiments, these hallucinations stem from the models' overconfidence in their relation predictions.
Specifically, LLMs tend to infer event relations without explicit linguistic cues or strong evidence for logical inference.
Consequently, LLMs predict far more relations than the gold standards, leading to low precision.

Recent studies show that LLMs can evaluate and correct their own outputs \cite{madaan2023selfrefine,asai2024selfrag}.
Thus, we introduce a hallucination grader to address hallucination.
For each relation edge generated, we pose a question to the LLMs to determine whether the relation is grounded in the given document.
If the LLMs respond with a ``yes'', the edge is retained; 
otherwise, it is discarded.
An example of the hallucination grader prompt is shown in Appendix (Table \ref{hallucination_prompt_table}).

\paragraph{Recover Missing Edges} The main benefit of the hallucination grader is that it increases precision by removing low-confident edges.
However, this process inevitably reduces recall.
To mitigate this side effect, we introduce an iterative refinement process.
After discarding hallucinated edges, we reinsert the code block containing the relation edges into the graph generation prompt and ask the LLMs to complete the code again.
In this way, the LLMs can reconsider whether there are any missing relations in the document, thereby improving recall.

Once the LLMs generate a new graph, the hallucination grader checks the relation edges again.
This process is repeated for a fixed number of times.
We set the maximum number of iterations to $5$ in our experiments, as the LLMs stop discovering new edges after $2$ or $3$ iterations in most documents.

\subsection{Complement Relation Types}

\label{progressive_gen}

Hierarchical, temporal, and causal relations are not independent of each other.
In this case, we found that providing the graph for the first relation can benefit the generation of the dependent relation's graph.
Specifically, we predict the hierarchical relation graph first.
Then, we provide this graph to the LLMs and ask them to generate the temporal relation graph.
Lastly, with both the hierarchical and temporal relation graphs available, the LLMs predict the causal relation graph.

The hierarchical relation describes two closely related events at different granularity levels.
It focuses on the inherent semantics of the events and does not depend on other relation types.
For example, ``\textit{writing a dissertation}'' is a subevent of ``\textit{doing a PhD}''. 
Therefore, we choose to predict the hierarchical relations first.
Temporal relations can depend on hierarchical relations.
For example, knowing ``\textit{doing a PhD}'' happened before ``\textit{being prompted to Professor}'' allows us to deduce that ``\textit{writing a thesis}'' also happened before ``\textit{being prompted to Professor}''. 
Thus, we predict temporal relations after hierarchical relations.
Lastly, causal relations depend on both hierarchical and temporal relations, as the antecedent event in a causal relation must occur before the consequence.
Therefore, the causal relation is predicted in the last step.
For more details about the entire prompting process, please refer to the pseudocode in Appendix \ref{sec:prompt_details}.

\subsection{Hungarian Graph Similarity}
\label{sec:hgs}

It is challenging to compare event relation graphs generated by LLMs due to the abstractive nature of generation, making it difficult to align the generated events with the gold standard events \cite{li2023evaluating}.
Moreover, salient events are often high-level and abstract rather than fine-grained and concrete, which means some variations in wording are not only acceptable but also expected.
Instead of using exact matching \cite{zhao2024large} or rule-based token matching \cite{tan-etal-2024-set} on events and relations to calculate $F_1$, adopting semantic-based evaluation metrics is more reasonable and fair.
As more tasks adopt text generation frameworks, many researchers are also turning to metrics based on language models rather than traditional token matching metrics like ROUGE and BLUE  \cite{goyal2022news,pratapa-etal-2023-background,lyu-pergola-2024-scigispy}.

In this study, we propose a novel metric for evaluating LLM-generated event graphs, called \textbf{Hungarian Graph Similarity (HGS)}. 
The metric is based on the Hungarian assignment algorithm \cite{kuhn1955hungarian}, which is widely used in object detection to match generated objects and target objects  \cite{carion2020end}.
It can find the optimal assignment given a cost matrix containing the distance between elements in two lists of objects.
We adapt this algorithm to match  predicted edges with edges in the gold standard graphs as follows:

\begin{enumerate}
    \item Encode the events using SFR-Embedding-Mistral \cite{SFRAIResearch2024}.
    \item Given two edges of the same relation type, let $\bar{e}^h_1,\bar{e}^t_1$ be the embeddings of the head event and the tail event in the first edge. Let $\bar{e}^h_2,\bar{e}^t_2$ be the embeddings of the head and tail events in the second edges. We define the distance between the edges as $\max\bigl(D_{cos}(\bar{e}^h_1, \bar{e}^h_2), D_{cos}(\bar{e}^t_1,\bar{e}^t_2)\bigr)$, where $D_{cos}(\cdot,\cdot)$ is the cosine distance.
    \item Build a cost matrix by computing the distance between every edge pair in the gold and predicted edge sets. Pad the matrix to a square matrix with the maximum cost value of $1$.
    \item Apply the Hungarian algorithm to the cost matrix to get the minimal cost value. The final score is $1-\mbox{cost value}$, making the value more intuitive (higher is better).
To compute the HGS over all the documents, we weight the scores by the number of gold edges to obtain an average value.
\end{enumerate}

In step $2$, we take the maximum value of the distances between head and tail events because relation edges are considered matched only if both the head and tail events match.

For more detailed analysis, we define precision-oriented HGS and recall-oriented HGS.
We match edges without padding the cost matrix in step $3$ to obtain the total cost values of all matched edge pairs.
Then, the total matched similarity is the number of matched edges minus the total cost.
\textbf{Precision-oriented HGS} is computed by dividing the total matched similarity by the total number of predicted edges.
\textbf{Recall-oriented HGS} is computed by dividing the total matched similarity by the total number of edges in the target graph. 

\section{Dataset}

In this section, we describe how we construct the \textit{NYT-SEG} which includes LLM-generated distant training pairs and a human-annotated dataset based on the New York Times (NYT) corpus.

\subsection{Document Selection}
We follow the same procedures as in \cite{tan-etal-2024-set} to select documents from the NYT corpus, one of the largest news datasets, with additional filtering based on document length. 
We select $10,347$ documents based on their descriptors indicating they are related to event narratives instead of opinions and discussions, such as sports and international politics.
Among them, $100$ documents are randomly sampled as the test set to be annotated by humans.
More details about data selection are shown in Appendix \ref{sec:dataset}.

\subsection{Annotation Settings}
\label{sec:annotation}
We recruited annotators from Prolific\footnote{\url{http://www.prolific.com}}.
There are two subtasks: \emph{salient event identification} and \emph{event relation identification}.
In the first subtask, the participants are asked to identify the salient event triplets: \emph{actor}, \emph{trigger}, and \emph{object} (optional).
We provide the definition of an event and several examples in the guidelines.
They are instructed to do the summarisation test: the salient events should be the events they would include in the summary of the given article.
Moreover, we provide some prominent features for helping annotators to identify salient events \cite{choubey-etal-2018-identifying,jindal-etal-2020-killed}:
\begin{itemize}
    \item Frequency: salient events are frequently mentioned in the articles.
    \item First appearance: salient events are often mentioned at the beginning of the article.
    \item Stretch size: salient events are often mentioned throughout the articles. Stretch size is the distance between the location where the event is first mentioned and last mentioned. A salient event usually has a large stretch size.
\end{itemize}

In the second stage, we ask participants to identify relation triplets: \emph{a source event}, \emph{a relation type}, and \emph{a target event}.
Both the source and target events should be among the salient events identified in the first stage.
In the guideline, we define three relation types: \textit{happened\_before}, \textit{caused\_by}, and \textit{is\_subevent\_of}.
\textit{happened\_before} indicates that the source event happened earlier than the target event.
\textit{caused\_by} means the source event would not have happened if the target event did not happen.
\textit{is\_subevent\_of} signifies that the source event is a subevent of the target event.
Annotators were informed that relations would be either explicitly mentioned in the article or inferred based on evidence within the article.
Further details about the guidelines and user interface can be found in Appendix \ref{sec:user_interface}.

\subsection{Inter Annotator Agreement}
Identifying salient events and event relations is complicated and time-consuming.
We found it challenging to educate participants about these concepts because, in daily life, the meanings of events and relations differ from their definitions in the field of information extraction.
Moreover, the technical definitions are much less intuitive to those outside the academic field.
As a result, thorough training of participants is important to obtain high-quality annotations.

In total, we recruited $3$ annotators to annotate $100$ documents.
Due to their varying availability, annotator $1$ and $2$ each annotated $45$ documents, while annotator $3$ annotated $20$ documents.
Among these, $5$ documents were annotated by all three annotators.
Following prior research in information extraction  \cite{gurulingappa2012development,zhao2024large}, we used $F_1$ to measure the inter-annotator agreement on these $5$ documents.
To compute inter-annotator agreement,  events or relations identified by one annotator are represented as set $S_1$.
Another annotator's annotation $S_2$ serves as a pseudo-reference to compute precision $=\frac{|S_1\cap S_2|}{|S_1|}$, recall $=\frac{|S_1\cap S_2|}{|S_2|}$, and the $F_1$ score $=\frac{2|S_1\cap S_2|}{|S_1| + |S_2|}$.

Table \ref{tab:iaa} shows the agreement scores for stages $1$ and $2$.
Identifying salient events is subjective,  which makes it difficult to reach a complete agreement.
Moreover, event relation identification is even more subjective and dependent on the previous stage, leading to less unanimous agreement. 

\begin{table}[h]
    \centering
    {\small
    \begin{tabular}{ccc}
      \toprule
       Annotator  & Stage $1$ & Stage $2$ \\
       \midrule 
       1 \& 2 & $0.838$ & $0.676$ \\
       1 \& 3 & $0.771$ & $0.645$ \\
       2 \& 3 & $0.847$ & $0.710$ \\
       Average & $0.819$ & $0.677$ \\
       \bottomrule
    \end{tabular}
    }
    \caption{Inter-annotator agreement measured by $F_1$.}
    \label{tab:iaa}
\end{table}

\subsection{Dataset Statistics}

Table \ref{tab:label_distribution} shows the distributions of the relation types after applying the transitive closure to the annotated graphs.
\textit{happened\_before} emerges as the most frequent relation type, reflecting the predominant focus on temporal sequences in news articles, and they are relatively straightforward to identify.
Conversely, \textit{caused\_by} is the least frequent as it is the most challenging to identify.

\begin{table}[h]
    \centering
    {\small
    \begin{tabular}{lc}
      \toprule
       Relation Type  & Number \\
       \midrule 
       happened\_before & $310$ \\
       caused\_by & $202$ \\
       is\_subevent\_of & $245$ \\
       Total & $757$ \\
       \bottomrule
    \end{tabular}}
    \caption{The distributions of the relation types.} \vspace{-15pt}
    \label{tab:label_distribution}
\end{table}

\section{Experiments}

\begin{table*}[ht]
  \begin{center}
  \begin{tabular}{lrrrr}
  \toprule
  \bf ~ & \multicolumn{1}{c}{Mean event number}  & \multicolumn{1}{c}{Event Frequency $\uparrow$}  & \multicolumn{1}{c}{First Appearance $\downarrow$}  & \multicolumn{1}{c}{Stretch Size $\uparrow$} \\
  \midrule 
  CAEVO & $34.71$ & $0.05$  & $0.46$ & $0.07$ \\
  Human & $8.26$ & $\bf{0.11}$ & $0.31$ & $\bf{0.20}$\\
  GPT-4 & $6.49$ & $0.09$ & $0.37$ & $0.18$ \\
    Llama3 & $5.17$ & $0.09$ & $\bf{0.30}$ & $0.19$ \\
      Mixtral & $10.60$ & $0.10$ & $0.33$ & $\bf{0.20}$ \\
  \bottomrule
  \end{tabular}
  \end{center}
  \caption{The average number of extracted events and the saliency features (in percentage values).}
  \label{tab:saliency_feature}
\end{table*}

\subsection{Model Settings}

We compare against the following baselines:
\begin{itemize}
    \item CAEVO \cite{mcdowell-etal-2017-event} is a pipeline system based on a MaxEnt classifier and manual features for extracting events and relations.
    \item \citet{madaan-yang-2021-neural} trained language models on CAEVO-generated linearised graphs with LM objective. We implemented their method to train a Flan-T5 model.
    \item \citet{tan-etal-2024-set} also trained language models on CAEVO-generated graphs, but applied data augmentations and regularisations to mitigate the set element misalignment issue. We applied their method to train a Flan-T5.
    \item \citet{han-etal-2019-joint} proposed a joint event and temporal relation extraction model. We adapted the model to hierarchical and causal relations by training on MAVEN-ERE. We also replaced BERT with Longformer \cite{DBLP:journals/corr/abs-2004-05150} to make it suitable for long documents.
    \item \textsc{GPT-4} \cite{openai2024gpt4} and \textsc{GPT-3.5} 
    are based on the generative pre-train framework. We used ``gpt-4-1106-preview'' and ``gpt-3.5-turbo'' respectively.
    \item \textsc{Mixtral} is an LLM based on the Mistral model and the mixture of expert framework. We used the 8x7B instruct  \cite{jiang2024mixtral}.
    \item \textsc{Llama3} is an LLM based on the Llama framework. 
    We used the 70B-instruct 8bit version for a balance between speed and performance.
\end{itemize}

We fine-tuned a Flan-T5-base ($250$M) with the relation graphs generated by CALLMSAE, following the same method as in \citet{tan-etal-2024-set}.
The baseline prompt evaluates whether each event pair is supported by the document, akin to the hallucination grader described in Section \ref{sec:grader}.
Thus, it serves as an ablation of our method without incorporating the code prompt.
Another baseline, which asks for the relation type given an event pair, is also tested and included in our experiment.

CALLMSAE is designed to be model-agnostic.
Due to budget constraints and the preliminary test results, we chose Llama3 as the backbone of all the prompt-based methods detailed in Table \ref{tab:overall_graph_eval}.

\subsection{Event Saliency Evaluation}

Table \ref{tab:saliency_feature} shows the salient features (defined in Section \ref{sec:annotation}, computation formulas in Appendix \ref{sec:features}) extracted from various backbone LLMs using summarisation prompts, alongside comparison with CAEVO and human annotations.
The LLM-generated events are much more salient than CAEVO-generated events and exhibit similarity to human annotations.

We also use human annotations to evaluate the saliency.
In the salient event identification annotation, we provide the events generated by CAEVO and Mixtral as candidate salient events.
Note that only the top CAEVO events ranked in saliency features are shown.
Half of the candidates are from CAEVO and the other half are from Mixtral.
They are randomly shuffled and then shown to the annotators.
We compute the precision, recall, and $F_1$ based on how the annotators select them.
We also compute HGS using human-annotated salient events as references (Table \ref{tab:saliency_human}). It is clear that although CAEVO extracted more events than Mixtral, many of them are not salient. Mixtral outperforms CAEVO significantly across all evaluation metrics.

\begin{table}[tb]
  \begin{center}
  {
  \begin{tabular}{lrrrr}
  \toprule
  \bf ~ & $P$ & $R$ & $F_1$ & HGS\\ 
  \midrule 
CAEVO & $3.29$& $3.72$  & $3.49$ & $18.18$\\
  Mixtral & $48.97$& $56.77$ & $52.59$ & $67.15$\\
  \bottomrule
  \end{tabular}}
  \end{center}
  \caption{Precision, recall, and $F_1$ based on the choices of the annotators. Hungarian graph similarity (HGS) is defined in Section \ref{sec:hgs}. The values are in percentage.}
  \label{tab:saliency_human}
\end{table}

\begin{table*}[tb]
  \begin{center}
  {\small
  \begin{tabular}{lrrrrrrrrr}
  \toprule
    & \multicolumn{3}{c}{Hierarchical}  & \multicolumn{3}{c}{Temporal}& \multicolumn{3}{c}{Causal}\\
   \cmidrule(lr){2-4}  \cmidrule(lr){5-7} \cmidrule(lr){8-10}
  \bf ~ & $PHGS$ & $RHGS$ & $HGS$ & $PHGS$ & $RHGS$ & $HGS$ & $PHGS$ & $RHGS$ & $HGS$\\ 
  \midrule
  \citet{han-etal-2019-joint}  & $0.158$ & $0.247$ & $0.098$ & $0.092$ & $0.352$ &$0.148$  & $0.084$ & $0.316$ & $0.116$\\
  CAEVO & - & - & - & $0.030$ & $0.558$ & $0.092$  & - & - & - \\
  \citet{madaan-yang-2021-neural}   & - & - & - & $0.061$ & $0.439$ & $0.116$ & - & - & -\\
  \citet{tan-etal-2024-set}  & - & - & - & $0.126$ & $0.335$ & $0.187$  & - & - & -\\
  \midrule
  Baseline Prompt  & $0.076$ & $\bf{0.651}$ & $0.248$  & $0.085$ & $0.627$ & $0.195$  & $0.062$ & $\bf{0.657}$ & $0.207$\\
  Baseline Prompt \tiny{(rel type)}  & $0.288$ & $0.375$ & $0.268$  & $0.135$ & $0.604$ & $0.261$  & $0.185$ & $0.513$ & $0.256$\\
  Code Prompt  & $0.174$ & $0.559$ & $0.315$ & $0.153$ & $\bf{0.678}$ & $0.283$  & $0.121$ & $0.632$ & $0.272$\\
  Code Prompt \tiny{(dependent rels)}   & N.A.& N.A.& N.A. & $0.211$ & $0.601$ & $0.341$  & $0.135$ & $0.599$ & $0.272$\\
  CALLMSAE \tiny{(ours)}  & $0.196$ & $0.544$ & $0.334$  & $\bf{0.294}$ & $0.509$ & $0.327$  & $0.198$ & $0.529$ & $0.295$\\
  Fine-tuned T5 \tiny{(CALLMSAE)} & $\bf{0.314}$ & $0.434$ & $\bf{0.339}$ & $0.244$ & $0.544$ & $\bf{0.362}$  & $\bf{0.366}$ & $0.397$ & $\bf{0.343}$\\
  \bottomrule
  \end{tabular}}
  \end{center}
  \caption{The Hungarian Graph Similarity (HGS) of the LLM-generated graphs on the human-annotated NYT dataset. PHGS is precision-oriented HGS. RHGS is recall-oriented HGS. \textit{Code Prompt (dependent rels)} means adding hierarchical graphs in the prompts for temporal graphs; and adding hierarchical and temporal for causal graphs. \textit{Fine-tuned T5 (CALLMSAE)} means fine-tuning a flan-T5 using the graphs generated by CALLMSAE. All prompt-based methods (row $5$ - $9$) are based on \textit{Llama3-70B-instruct}.
  \label{tab:overall_graph_eval}}
\end{table*}

\subsection{Salient Event Relation Graph Evaluation}

The salient event relation graph evaluation results are shown in Table \ref{tab:overall_graph_eval}.
The compared methods (row $1$ - $4$) are outperformed by \textit{Baseline Prompt} (row $5$) on all relation types.
\textit{Baseline Prompt (rel type)} (row $6$), which asks the model to generate all possible relation types given the event pair, performs slightly better than \textit{Baseline Prompt}.
However, \textit{Baseline Prompt} and \textit{Baseline Prompt (relation type)} are slow and costly because the number of prompts they need for building one graph is $O(n^2)$, where $n$ is the number of events in the document.
On the other hand, the time complexity of \textit{Code Prompt} is $O(1)$.
Moreover, \textit{Code Prompt}'s overall HGS is significantly higher than \textit{Baseline Prompt} and \textit{Baseline Prompt (relation type)} on all relation types.
\textit{Baseline Prompt} check the event pairs more thoroughly and thus have higher recall but its precision is much lower.
The complete CALLMSAE combines the code prompt and hallucination grader for iterative refinement, checking missing relations and verifying them to prevent hallucination.
\textit{Code Prompt (dependent rels)} (the $8$th row) is an ablation of \textit{CALLMSAE} (the $9$ row), differing only in the absence of iterative refinement. 
These results highlight the effectiveness of the hallucination grading approach, which effectively increases the precision and strikes a balance with recall. 

In the temporal category, the results of \emph{Code Prompt \small{(dependent rels)}} are obtained when provided with hierarchical graphs generated by CALLMSAE to LLMs.
It has much higher overall HGS and precision than \textit{Code Prompt} without hierarchical information, showing that hierarchical information can mitigate hallucinations during the temporal graph generation.
In the casual category, the results of \emph{Code Prompt \small{(dependent rels)}} are obtained when given both hierarchical and temporal graphs generated by CALLMSAE.
The additional information also increases precision.

\textit{Fine-tuned T5} outperform all the methods based on CAEVO \cite{mcdowell-etal-2017-event,madaan-yang-2021-neural,tan-etal-2024-set}, showing that the high-quality graphs generated by CALLMSAE can boost the contextualised graph generation.
Interestingly, the performance of the \textit{Fine-tuned T5}, fine-tuned on CALLMSAE-generated data, exceeds that of CALLMSAE itself, implying that the fine-tuned model can effectively adapt the reasoning patterns provided by Llama3 and generalise them.

\begin{table}[htb]
  \begin{center}
  {\small
  \begin{tabular}{lrrrr}
  \toprule
  \bf ~ & Hier & Temp & Causal & Overall  \\ 
  \midrule 
 Baseline (rel type) & $0.58$& $0.66$ & $0.48$ & $0.60$ \\
  CALLMSAE & $0.71$& $0.71$ & $\bf{0.65}$ & $0.69$ \\
  Fine-tuned T5 & $\bf{0.72}$& $\bf{0.74}$ & $0.62$ & $\bf{0.70}$  \\
  \bottomrule
  \end{tabular}}
  \end{center}
  \caption{The human evaluation scores.} 
  \label{tab:human_eval}
\end{table}

\subsection{Human Evaluation of Event Graph}
We recruited additional annotators to evaluate the generated graph on $50$ of the test documents.
We asked them whether the relation edges are correct.
These scores ($\frac{\textrm{correct edges}}{\textrm{generated edges}}$) can be viewed as precision (Table \ref{tab:human_eval}).
The human evaluation correlates with the HGS scores, verifying our conclusion.

\begin{table}[htb]
  \begin{center}
  {\small
  \begin{tabular}{lrr}
  \toprule
  \bf ~ & Format Error $\downarrow$ & Cycle $\downarrow$ \\ 
  \midrule 
 GPT-3.5& $0$\%& $10.67$\%   \\
  GPT-4 & $3.67$\%& $1.67$\%  \\
  Mixtral & $3.33$\%& $2.33$\%  \\
  Llama3 & $\bf{0}$\%& $\bf{0}$\%  \\
  \bottomrule
  \end{tabular}}
  \end{center}
  \caption{The average number of CALLMSAE-generated graphs out of $100$ with format errors or cycles.}
  \label{tab:format}
\end{table}

\subsection{Format Error and Cycles in the Graphs}

We specified the relation graphs as directed acyclic graphs in the prompt.
If there is a cycle in the generated graph, it means that the LLM failed to follow the instructions.
A cycle also indicates constraint violations because all the relations in the graphs are asymmetric.
We use the APIs in \textit{NetworkX} package to detect cycles in the transitive closure of the graphs.
If the Python interpreter returns an error, it is classified as a format error.
We prompt each LLM three times on the test set.
Table \ref{tab:format} shows the average number of documents encountering format errors or cycles.
All LLMs have low rates of format errors which shows that state-of-the-art LLMs can understand the instruction well and generate executable Python code.
Among them, GPT-3.5 and Llama3 have no errors.
About $10\%$ of graphs generated by GPT-3.5  have cycles, suggesting that GPT-3.5 may have inferior reasoning ability compared to other LLMs.
GPT-4 and Mixtral both have low rates of cycle occurrence, but they are beaten by Llama3 which has no cycle in all generations, showing its remarkable understanding of the transitive and asymmetric constraints in the complex event relation graphs.

\section{Conclusion}

This study explored utilising LLMs to generate salient event relation graphs from news documents without relying on human annotations.
We studied how the events generated by LLMs are compared to the traditional methods in terms of event saliency.
We further demonstrated that CALLMSAE-generated graphs can serve as distant signals to fine-tune smaller models and outperform those based on CAEVO. 
The CALLMSAE-generated event graphs, together with the human-annotated test set are collected as \textit{NYT-SEG}.

\section*{Limitations}

CALLMSAE is more demanding than CAEVO in terms of computational power and time.
To generate the \textit{NYT-SEG}, we spent $2,200$ wall clock hours in inference.
More details about resource cost are disclosed in Appendix \ref{sec:prompt_details}.

Although we have tested many prompting methods and included several of the most effective ones in this paper, we have not explored all possible combinations due to the extensive volume of recent literature on prompt engineering.
There might still exist combinations of prompts that could further improve performance~\cite{alzaid24,zhou-etal-2024-mystery}.
However, we are almost certain that any potential combinations, if they exist, are likely to be more complex and thus less efficient for building large-scale datasets.
For example, we did not add demonstrations in graph generation because the code template is already quite lengthy. Adding more documents could potentially exceed the context windows of some LLMs, making it challenging for them to interpret the instructions effectively. 
The main goal of this work is to demonstrate the potential of LLM-based generation can help the data-demanding event graph generation task. 

\section*{Ethics Statement}

Event relation graph generation is a powerful tool for understanding text.
A potential misuse of the proposed method is mining user behaviours on their private data.
For example, salient event relation graphs can be extracted from users' tweets to analyse their potential reactions to advertisements and scams.
That could be a huge risk to social media users.

Another potential risk is that the saliency may introduce bias.
LLMs may have their preferences in selecting a specific group of events as important events due to the data they were trained on.
This is a question which requires further large-scale investigation.
However, we think this risk is negligible in this study because we work on document-level information.
There is little room for selection given that the news articles are already the products of choice and distillation.
If the system is used to extract information from a border information source, such as social media, the risk must be carefully assessed.

\section*{Acknowledgements}
This work was supported in part by the UK Engineering and Physical Sciences Research Council (EPSRC) through a Turing AI Fellowship (grant no. EP/V020579/1, EP/V020579/2).
Xingwei Tan was supported by the Warwick Chancellor's International Scholarship.
This work was conducted on the Sulis Tier-2 HPC platform hosted by the Scientific Computing Research Technology Platform at the University of Warwick. Sulis is funded by EPSRC Grant EP/T022108/1 and the HPC Midlands+ consortium.

\bibliography{custom.bib, anthology}.bib
\bibliographystyle{acl_natbib}

\newpage
\appendix

\section{Additional Details of Dataset Construction}

\subsection{Document Selection}
\label{sec:dataset}
We select news documents from the NYT corpus based on the descriptors available.
With regards to the generation of salient even graphs, the most relevant documents tend to be centered around event narratives, so that they could be rich in event relations.
\citet{tan-etal-2024-set} investigated which descriptors are rich in event narrative using event frequency $\times$ inverse-descriptor frequency.
We chose the documents using the same descriptors as them (e.g., ``\textit{airlines and airplanes}'', ``\textit{united states international relations}'',
                      ``\textit{civil war and guerrilla warfare}'',
                      ``\textit{track and field}'', ``\textit{soccer}'', etc.).

We applied additional filtering based on the number of words in the documents.
Documents with more than $8500$ words or less than $100$ words are excluded.
Based on our preliminary observations, the extremely long documents are not typically news articles (only takes $0.02\%$ in the entire NYT).
They tend to be collections of articles over longer time spans, making them not suitable as focus of this study.
Additionally, very long articles may affect the performance of open-source LLMs only due to limitations in the context length rather than their reasoning abilities. 
On the other hand, articles that are too short are less likely to contain complex event relation graphs, so we also exclude them.
The final average word count of the selected $10347$ documents is $780$.

\subsection{Frequent words and descriptors in the annotated dataset}
\label{sec:frequent_words_descriptors}

\begin{table}[ht]
    \centering
    \begin{tabular}{lcccc}
      \toprule
      & \multicolumn{2}{c}{Test}  & \multicolumn{2}{c}{Train} \\
   \cmidrule(lr){2-3}  \cmidrule(lr){4-5}
       Rank  & Word & Count & Word & Count\\
       \midrule 
       1 & win & $41$ & win & $2,964$\\
       2 & express & $15$ & make & $1,591$\\
       3 & play & $14$ & face & $1,564$\\
       4 & make & $13$ & express & $1,411$\\
       5 & defeat & $12$ & include & $1,307$\\
       \bottomrule
    \end{tabular}
    \caption{The top $5$ most frequent trigger words in the human-annotated test set and the distant train set.}
    \label{tab:frequent_triggers}
\end{table}

\begin{table*}[ht]
    \centering
    {\small
    \begin{tabular}{lcccc}
      \toprule
      & \multicolumn{2}{c}{Test}  & \multicolumn{2}{c}{Train} \\
   \cmidrule(lr){2-3}  \cmidrule(lr){4-5} 
       Rank  & Descriptor & Count & Descriptor & Count\\
       \midrule 
       1 & U.S. International Relations & $27$ & Terrorism & $2,885$\\
       2 & Terrorism & $21$ & U.S. International Relations & $2,574$\\
       3 & Bombs and Explosives & $17$ & Bombs and Explosives & $1,727$\\
       4 & U.S. Armament and Defense & $15$ & U.S. Armament and Defense & $1,717$\\
       5 & Politics and Government & $15$ & Politics and Government & $1,649$\\
       \bottomrule
    \end{tabular}}
    \caption{The top $5$ most frequent descriptors in the human-annotated test set and the distant train set.}
    \label{tab:frequent_descriptors}
\end{table*}

Table \ref{tab:frequent_triggers} reports the most frequent trigger words among the human-identified salient events and LLM-generated salient events after filtering out the light words (words that have no semantic meaning).
We could see that ``\textit{win}'', ``\textit{play}'', and ``\textit{defeat}'' are prominent triggers due to the sports topics within the dataset.
These articles usually mention multiple events with these triggers.
Triggers like ``\textit{express}'', ``\textit{include}'', and ``\textit{make}'' are instead common across different scenarios.

Table \ref{tab:frequent_descriptors} shows the most frequent descriptors in the human-annotated test set and the distant train set.
These are the typical event-rich topics and are full of narratives.

\subsection{Disclaimers of Risks}
\begin{figure}[hb]
    \centering
    \includegraphics[width=0.9\columnwidth]{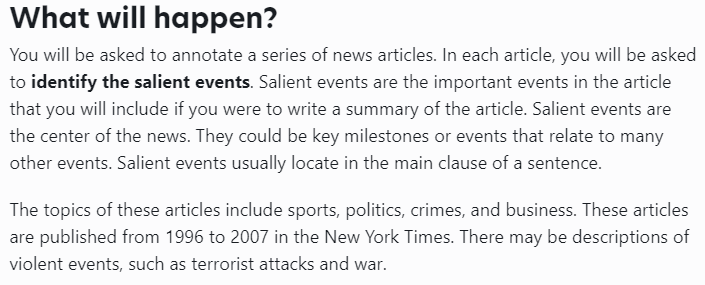}
    \caption{The recruitment descriptions.}
    \label{fig:recruitment}
\end{figure}

Consider that a large portion of the new articles in the New York Times corpus are about violent incidences, such as terrorist attacks and war.
To prevent inflicting harm to traumatised victims, we show the information clearly in the recruitment description on the Prolific platform (Figure \ref{fig:recruitment}).

\subsection{Guidelines and User Interface}
\label{sec:user_interface}
A well-designed user interface is essential for collecting high-quality data efficiently.
We fully cooperate with participants to improve the user interface iteratively based on their feedback.

\begin{figure*}
    \centering
    \includegraphics[width=0.9\linewidth]{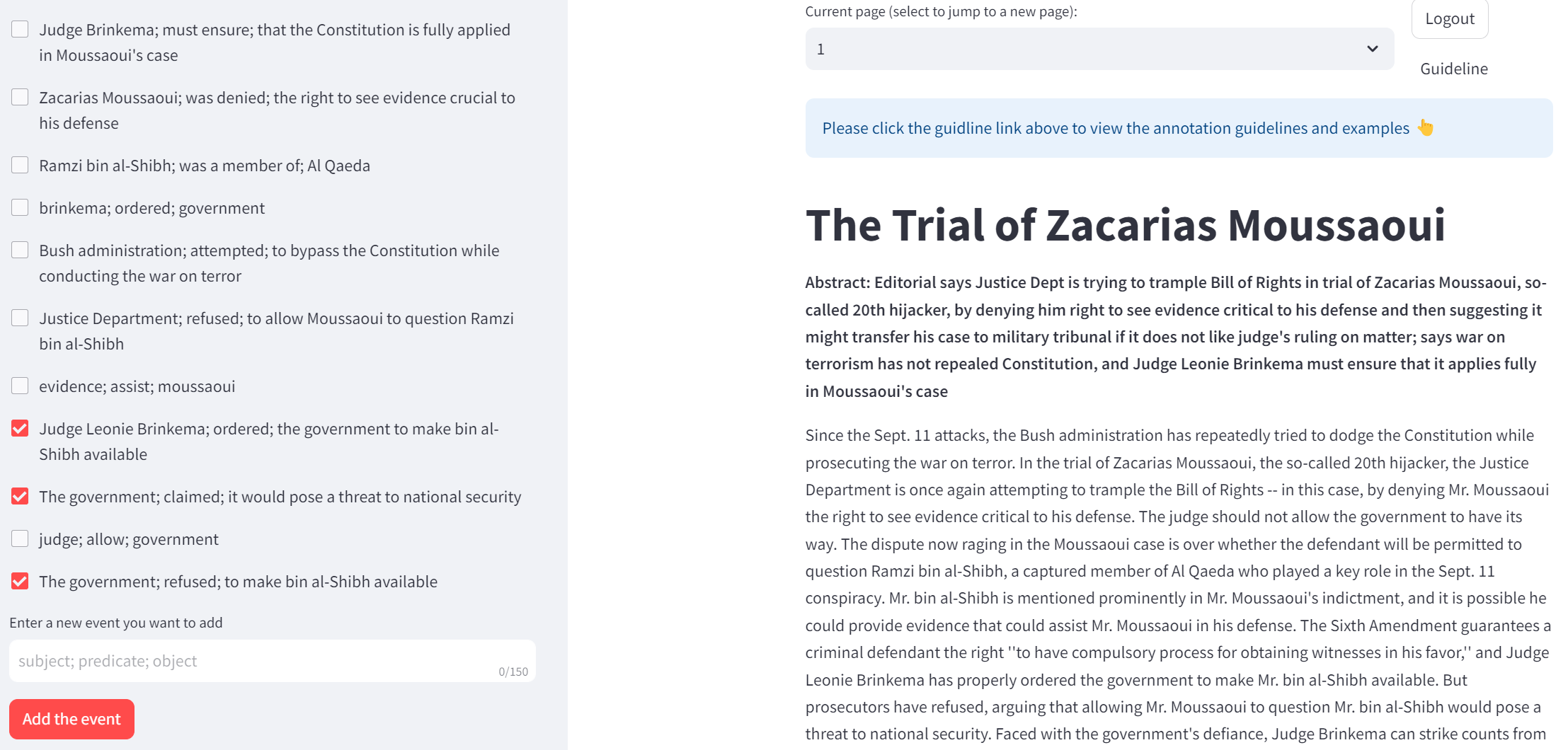}
    \caption{The user interface of salient event identification.}
    \label{fig:stage1}
\end{figure*}

In the salient event identification stage (Figure \ref{fig:stage1}), we show the title, abstract, and content of the article on the right side.
We show candidate events, which are extracted through CAEVO and Mixtral, on the left sidebar.
The shown CAEVO events are the top events ranked based on the saliency feature score.
The participants can choose the candidate events which they think are accurate and salient.
The guideline also informs them that if multiple options refer to the same event, they can only choose the most accurate and informative one.
If a salient event is not present among the candidates, they could write it in the text input box and add it.

\begin{figure*}
    \centering
    \includegraphics[width=0.9\linewidth]{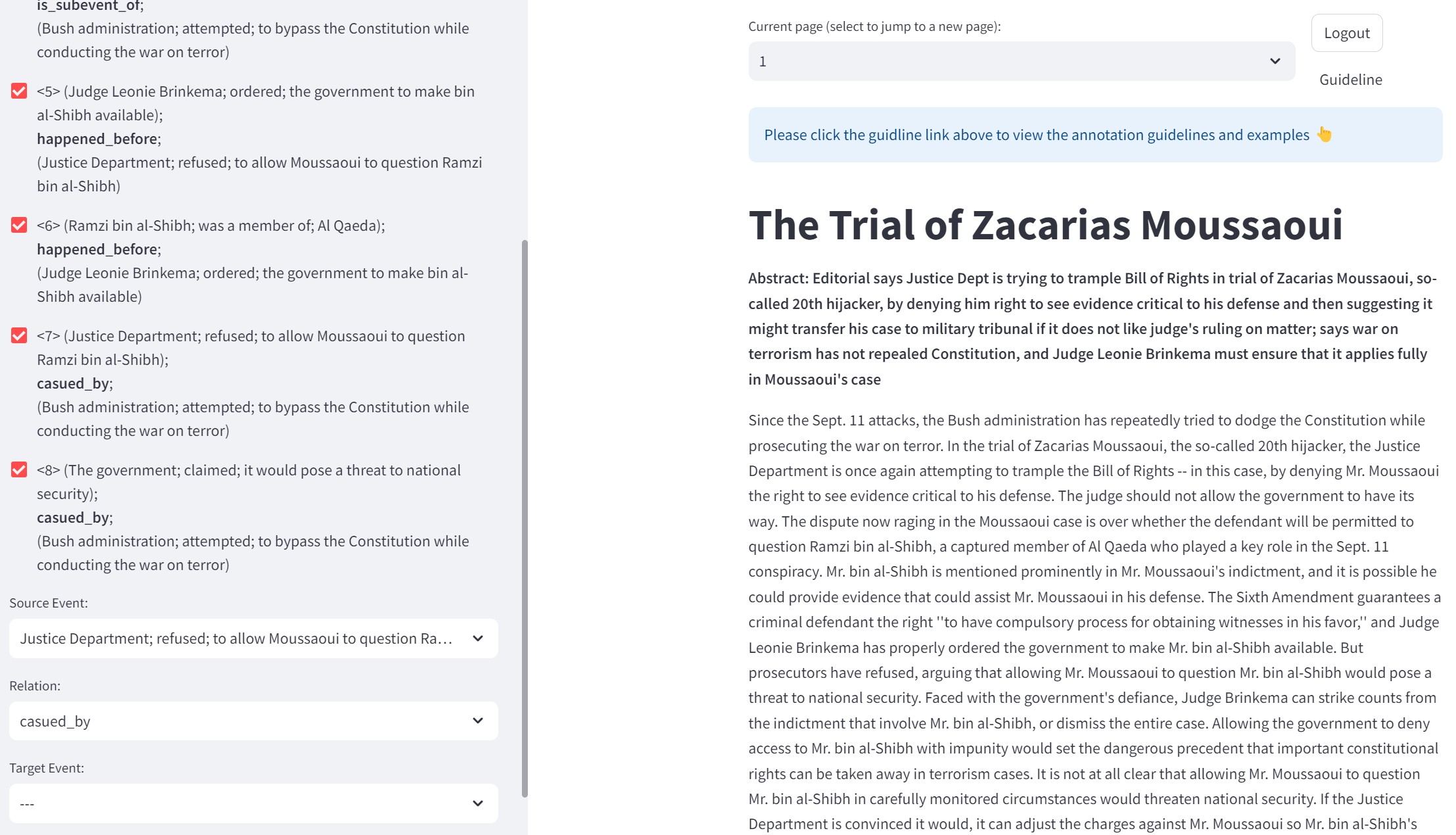}
    \caption{The user interface of event relation identification.}
    \label{fig:stage2}
\end{figure*}

In the event relation identification stage (Figure \ref{fig:stage2}), they could choose a source event, a relation type, and a target event to add a relation triplet.
The source event and the target event need to be chosen from the salient event list from the first stage.
We automatically detect and prevent any new event that will lead to duplication and contradiction.
The participants can also deselect the added event if they change their minds.
The participants were asked to finish the first stage first, and then annotate the second stage based on their own annotations in the first stage.

\begin{figure*}
    \centering
    \includegraphics[width=0.7\linewidth]{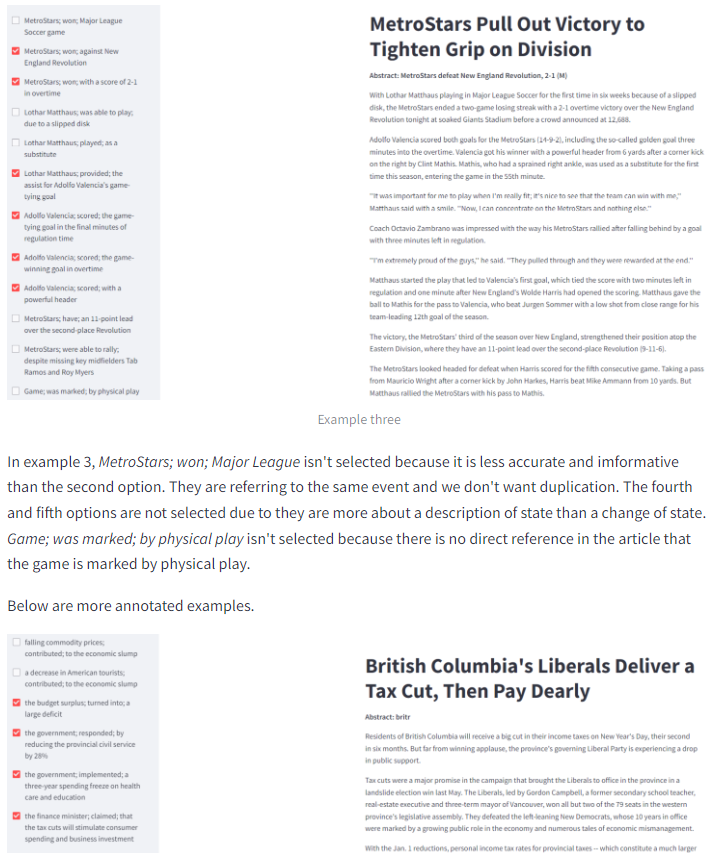}
    \caption{Annotation guidelines of salient event identification shown to the annotators.}
    \label{fig:guidelines_screen}
\end{figure*}

In the following are reported the screenshots of the guideline pages (Figure \ref{fig:guidelines_screen}).

\subsection{More details about the annotation}

We started the annotation process by releasing several trial rounds, during which we chose participants based on their dedication and understanding of the terminologies. It required considerable communication efforts to ensure they had an accurate understanding of the task definition.

During training, we found a common mistake among the annotators was that they tended to overestimate the \textit{is\_subevent\_of} relation. They often confused it with the \textit{caused\_by} relation or temporal inclusion.

We advised them that \textit{is\_subevent\_of} pertains to two events on different granularity levels but referring to the same subject. To distinguish \textit{is\_subevent\_of} from temporal overlap, they could check whether the actor in the subevent is the same as or a part of the actor or object in the parent event. For example, if a parent event is ``\textit{a team did something}'' the subevent can be ``\textit{a member of the team did something}''.

\subsection{Information about the Annotators}

The annotators were paid at the rate of $8$£/h.
We screened native English speakers from all over the world to ensure they could read English articles fluently. We also selected participants based on their previous submissions and approval rates to ensure they were familiar with the platform and were high-quality annotators. 

Two of the final annotators are identified as male, and they both come from the UK. One of the final annotators is identified as female, and she comes from Canada. They all identified as white.

\subsection{Dataset Licensing}

The original NYT corpus is available for noncommercial research license.
One of our authors has obtained the license.
Based on the license, we could not include the original text in our dataset.
Thus, we will only release the generated/annotated graphs.
Our dataset will also be in noncommercial research license.

\section{Saliency Features}
\label{sec:features}

Inspired by \cite{choubey-etal-2018-identifying}, we calculate the saliency features to show how our proposed method differs from previous methods in terms of event saliency.
Unlike conventional computation methods, these saliency features are calculated on the sentence level to be comparable across documents of various lengths.
These saliency features are:

\noindent \textbf{Event frequency}: A salient event tends to appear frequently in the document. 
    Let $D=\{s_0,s_1,...,s_{n-1},s_n\}$ be the document and the list of sentences in the document.
    Let $e$ be the event.
    Let $M(e)=\{s_i,s_j,...,s_k\},0\leqslant i<j<k\leqslant n$ be the list of sentences which mention the event $e$.
    The event frequency is calculated as:

\begin{equation}
    frequency(D,e) = \frac{|M(e)|}{n+1}.
\end{equation}
    
\noindent \textbf{First appearance}: News writers usually mention the salient event as early as possible to attract readers' attention. 
    The first appearance of the event $e$ is computed as:
    \begin{equation}
        first\_ appearance(D,e) = \frac{i}{n}.
    \end{equation}
    
\noindent \textbf{Stretch size}: Salient events tend to be mentioned all across the document.
    The stretch size of event $e$ is calculated as:
    \begin{equation}
        stretch\_size(D,e) = \frac{k-i}{n}.
    \end{equation}

To detect which sentences mention the event $e$, we first lemmatise the words in the document and the given event.
Then, detect whether there is a matched substring the same as the given event in each sentence.
However, the abstractive nature of LLM-based salient event generation makes exact matching not viable.
To detect the event mention of LLM-generated events, we formulate a series of prompts.
We first ask: ``\textit{Which sentence in the document below mentions the event "\{event\}"? Please enclose that sentence in () and show it. Document: """\{doc\textunderscore content\}}"""''.
Then, we employ iterative refinement in case the LLM misses any other sentences: ``\textit{Is there any other sentence in the document directly mentioning the event "\{event\}"? Please enclose that sentence in () and show it.}''
Lastly, we collect the sentences from the responses.

We run the methods on the human-annotated dataset ($100$ documents).
We compute the saliency features of the events in each document and take the average across the events.
Lastly, all the values are averaged across all the documents.

\begin{table}[htb]
  \begin{center}
  \begin{tabular}{lrrr}
  \toprule
  \bf ~ & Hier & Temp & Causal \\ 
  \midrule 
  JSON & $0.310$& $0.277$ & $0.251$  \\
  Python & $\bf{0.315}$& $\bf{0.283}$ & $\bf{0.272}$  \\
  \bottomrule
  \end{tabular}
  \end{center}
  \caption{The comparison between Python format prompt and JSON format prompt in Hungarian Graph Similarity.}
  \label{tab:fail_attempt}
\end{table}

\definecolor{rowcolor}{gray}{0.9}
\definecolor{linecolor}{gray}{0.6}

\setlength{\arrayrulewidth}{1pt}
\arrayrulecolor{linecolor}

\begin{table*}
  \begin{center}
  \resizebox{\textwidth}{!}{%
  \begin{tabularx}{\textwidth}{|X|X|}
\hline
\rowcolor{rowcolor} Input to the LLMs & Response from the LLMs \\ 
  \hline

  Write a summary of the document below using one paragraph. 

Document: """
Residents of British Columbia will receive a big cut in their income taxes on New Year's Day, their second in six months. But far from winning applause, the province's governing Liberal Party is experiencing a drop in public support.

Tax cuts were a major promise in the campaign that brought the Liberals to office in the province in a landslide election win last May. The Liberals, led by Gordon Campbell, a former secondary school teacher, real-estate executive and three-term mayor of Vancouver, won all but two of the 79 seats in the western province's legislative assembly. They defeated the left-leaning New Democrats, whose 10 years in office were marked by a growing public role in the economy and numerous tales of economic mismanagement.

With the Jan. 1 reductions, personal income tax rates for provincial taxes -- which constitute a much larger share of overall taxes for Canadians than state taxes do for Americans -- will have been chopped by an average of 25 percent, turning British Columbia from one of Canada's most heavily taxed provinces to one with some of the lowest tax rates.

...

"""

Summary:

  &

The Liberal Party government in British Columbia, led by Gordon Campbell, has implemented a significant cut in income taxes on New Year's Day, fulfilling a campaign promise, but it has coincided with an economic downturn. Despite the tax cuts, the province is facing a large deficit due to falling tax revenues and increased demand for social services. To address this, the government plans to reduce the civil service by 28\%, impose a three-year spending freeze on healthcare and education, and tighten benefits under the government-financed drug plan. However, many British Columbians are unhappy with these measures, and the Liberal Party's approval rating has dropped significantly, amidst concerns about the return of political polarization in the province.

\\
  \hline
  \end{tabularx}}
  \end{center}
  \caption{An example of the hallucination prompt input and the response from Llama$3$.}
  \label{summary_prompt_table}
\end{table*}

\begin{table*}
  \begin{center}
  \resizebox{\textwidth}{!}{%
  \begin{tabularx}{\textwidth}{|X|X|}
\hline
\rowcolor{rowcolor} Input to the LLMs & Response from the LLMs \\ 
  \hline
  
A structured event is something that happened as described in the text. A structured event is represented as a tuple, which consists of actors, a trigger, and objects. Could you list all the structured events in the following article? Example: 1. (John; married; Alice). 2. (Alice; was hired; by Google). 

Article: """{summary}"""

  &

Here are the structured events extracted from the article:

1. (Gordon Campbell; implemented; a significant cut in income taxes)

2. (The Liberal Party government; fulfilled; a campaign promise)

3. (The government; plans to reduce; the civil service by 28\%)

4. (The government; will impose; a three-year spending freeze on healthcare and education)

5. (The government; will tighten; benefits under the government-financed drug plan)

6. (Many British Columbians; are unhappy with; these measures)

7. (The Liberal Party's approval rating; has dropped significantly)

Note that some of these events may be implicit or inferred from the text, rather than explicitly stated.

\\
  \hline
  \end{tabularx}}
  \end{center}
  \caption{An example of the salient event generation prompt input and the response from Llama$3$.}
  \label{event_prompt_table}
\end{table*}

\begin{table*}
  \begin{center}
  \resizebox{\textwidth}{!}{%
  \begin{tabularx}{\textwidth}{|X|X|}
\hline
\rowcolor{rowcolor} Input to the LLMs & Response from the LLMs \\ 
  \hline
Could you please finish the following code?

import networkx as nx

document = \{document\}

event\textunderscore list = \{event\textunderscore list\textunderscore str\}

\# This is a graph representing the hierarchical relation between the events in the document

\# Each edge in the graph represents a subevent relation between the head and tail nodes which are events

\# An edge means the head event is a subevent of the tail event. They are closely related but on different granularity levels.

hierarchical\textunderscore graph = nx.DiGraph() \# This is a directed acyclic graph. There should not be any cycle in the graph.

\# Add events as nodes

for event in event\textunderscore list:

\setlength{\parindent}{1cm}    hierarchical\textunderscore graph.add\textunderscore node(event)

\noindent\# Add hierarchical relations as edges to the graph using .add\textunderscore edge() function. The function takes two strings as inputs. Each string represents an event.

\noindent\# There should not be any cycle in the graph. Explain the reason for each added edge as a comment after each function call
    
  &

Here is the completed code:

import networkx as nx

hierarchical\textunderscore graph = nx.DiGraph()

\# Add events as nodes

for event in event\textunderscore list:

\setlength{\parindent}{1cm}    hierarchical\textunderscore graph.add\textunderscore node(event)

\noindent\# Add hierarchical relations as edges to the graph

\noindent hierarchical\textunderscore graph.add\textunderscore edge(``the government responded by reducing the provincial civil service by 28\%'', ``liberals cut spending'') 

\noindent\# The event ``the government responded by reducing the provincial civil service by 28\%'' is a subevent of ``liberals cut spending'' because it is one way the Liberals cut spending.


\\
  \hline
  \end{tabularx}}
  \end{center}
  \caption{An example of the relation graph generation prompt input and the response from Llama$3$.}
  \label{prompt_table}
\end{table*}


\begin{table*}
  \begin{center}
  \resizebox{\textwidth}{!}{%
  \begin{tabularx}{\textwidth}{|X|X|}
\hline
\rowcolor{rowcolor} Input to the LLMs & Response from the LLMs \\ 
  \hline
  
You are a grader assessing whether an answer is grounded in / supported by a set of facts. Give a binary score `yes' or `no' score to indicate whether the answer is grounded in or supported by a set of facts. Then, provide a short explanation.

Here are the facts: \{document\}

Here is the answer: Event ``the government responded by reducing the provincial civil service by 28\%'' is a subevent of event ``liberals cut spending''.
  &

Score: Yes

Explanation: The answer is grounded in the facts because it accurately identifies a specific action taken by the government (reducing the provincial civil service by 28\%) as a subevent of the broader event of cutting spending, which is mentioned in the text.


\\
  \hline
  \end{tabularx}}
  \end{center}
  \caption{An example of the hallucination prompt input and the response from Llama$3$.}
  \label{hallucination_prompt_table}
\end{table*}

\section{Comparison of Formats}

In our preliminary experiments, we tested asking LLMs to generate event relations in JSON format (Table \ref{tab:fail_attempt}).
When generating JSON response, we found LLMs often make format errors (e.g., using double quotation marks in the wrong place) which makes the responses difficult to parse, resulting in lower HGS.
We also found the JSON format is not as flexible as the Python format which can integrate more information (e.g., explanations and definitions) as comments.

\section{Prompting Details}
\label{sec:prompt_details}
Due to limited resources, the sampling of the distant supervision dataset was conducted on multiple machines with different specifications, including one with $6\times$ RTX $3090$, one with an A$100$, and one with $2\times$ A$40$.
The total time cost for prompting Llama3 to construct the training data ($10,231$ documents) is about $2,200$ hours (the wall-clock time of all the machines combined).

Table \ref{prompt_table} shows an example of the code prompt for hierarchical graph generation and the response from Llama3.
Table \ref{hallucination_prompt_table} shows an example of the hallucination prompt and the response.

Algorithm \ref{alg:main} is the pseudo-code of the entire salient event graph generation process.

In the summarization prompt, we use a temperature of $0.8$ and a top\_p of $0.9$. For the salient event generation prompt, we use a temperature of $0.5$ and a top\_p of $0.9$. The relation graph generation prompt also uses a temperature of $0.5$ and a top\_p of $0.9$. The hallucination grader prompt uses a temperature of $0$.

\begin{algorithm*}
\caption{CALLMSAE: CAscading Large Language Models for SAlient Event graph generation}\label{alg:main}
\hspace*{\algorithmicindent}\textbf{Input:} Document $d$, Max Refinement Round $k$\\
\hspace*{\algorithmicindent}\textbf{Output:} An Event Relation Graph $g$
\begin{algorithmic}[1]
\State $summary \gets \textrm{Summary\_Generation}(d)$
\State $salient\_events \gets \textrm{Event\_Generation}(summary)$
\State $hierarchical\_graph \gets \textrm{null}$
\State $current\_round \gets 0$
\While{$current\_round < n$}
    \State $hierarchical\_graph \gets \textrm{Hierarchical\_Graph\_Generation}(d,salient\_events,$ \par $hierarchical\_graph)$
    \State $hierarchical\_edges \gets \textrm{Get\_Edges}(hierarchical\_graph)$ 
    \For{$edge_i$ in $hierarchical\_edges$}
        \State $remove\_edge \gets \textrm{Hallucination\_Grader}(d, edge_i)$ 
        \If{$remove\_edge$}
            \State $hierarchical\_graph \gets \textrm{Remove\_edge}(hierarchical\_graph,edge_i)$
        \EndIf
    \EndFor
    \State $current\_round \gets current\_round+1$
\EndWhile
\State $temporal\_graph \gets \textrm{null}$
\State $current\_round \gets 0$
\While{$current\_round < n$}
    \State $temporal\_graph \gets \textrm{Temporal\_Graph\_Generation}(d,salient\_events,temporal\_graph,$ \par $hierarchical\_graph)$
    \State $temporal\_edges \gets \textrm{Get\_Edges}(temporal\_graph)$
    \For{$edge_i$ in $temporal\_edges$}
        \State $remove\_edge \gets \textrm{Hallucination\_Grader}(d, edge_i)$ 
        \If{$remove\_edge$}
            \State $temporal\_graph \gets \textrm{Remove\_edge}(temporal\_graph,edge_i)$
        \EndIf
    \EndFor
    \State $current\_round \gets current\_round+1$
\EndWhile
\State $causal\_graph \gets \textrm{null}$
\State $current\_round \gets 0$
\While{$current\_round < n$}
    \State $causal\_graph \gets \textrm{Causal\_Graph\_Generation}(d,salient\_events,causal\_graph,$ \par $temporal\_graph,hierarchical\_graph)$
    \State $causal\_edges \gets \textrm{Get\_Edges}(causal\_graph)$
    \For{$edge_i$ in $causal\_edges$}
        \State $remove\_edge \gets \textrm{Hallucination\_Grader}(d, edge_i)$ 
        \If{$remove\_edge$}
            \State $causal\_graph \gets \textrm{Remove\_edge}(causal\_graph,edge_i)$
        \EndIf
    \EndFor
    \State $current\_round \gets current\_round+1$
\EndWhile
\State $g \gets \{hierarchical\_graph,temporal\_graph,causal\_graph\}$
\end{algorithmic}
\end{algorithm*}

\begin{figure*}
    \centering
    \includegraphics[width=0.7\linewidth]{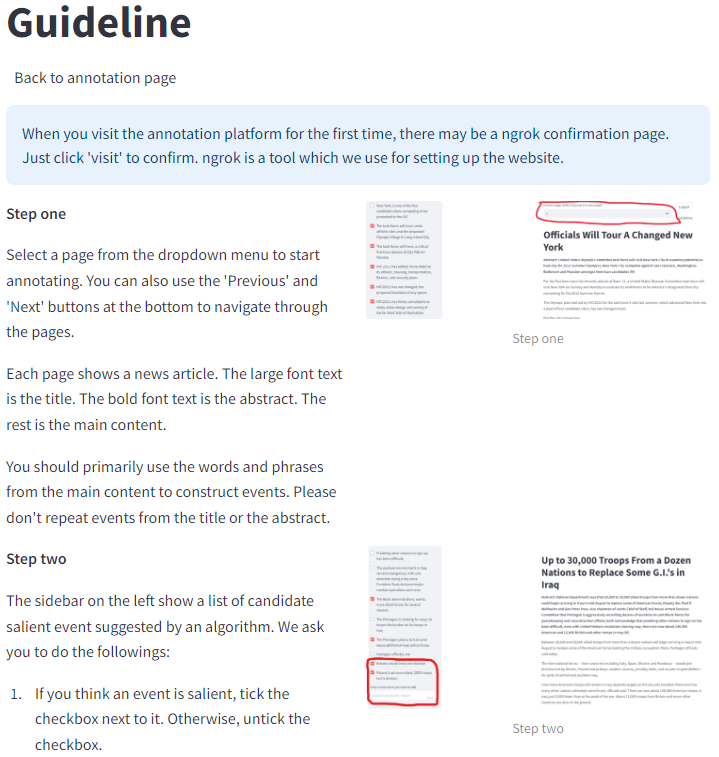}
\end{figure*}

\begin{figure*}
    \centering
    \includegraphics[width=0.7\linewidth]{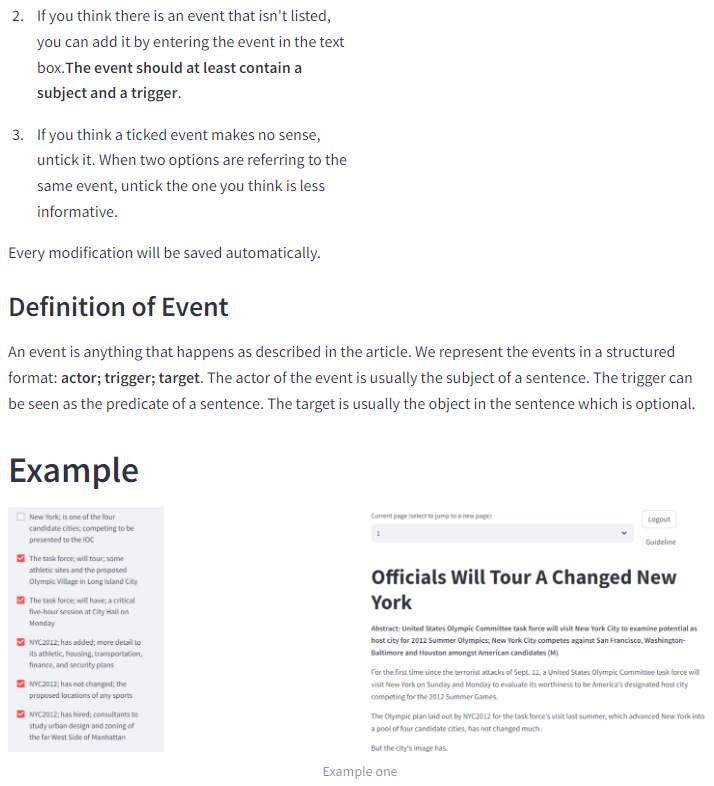}
\end{figure*}

\begin{figure*}
    \centering
    \includegraphics[width=0.7\linewidth]{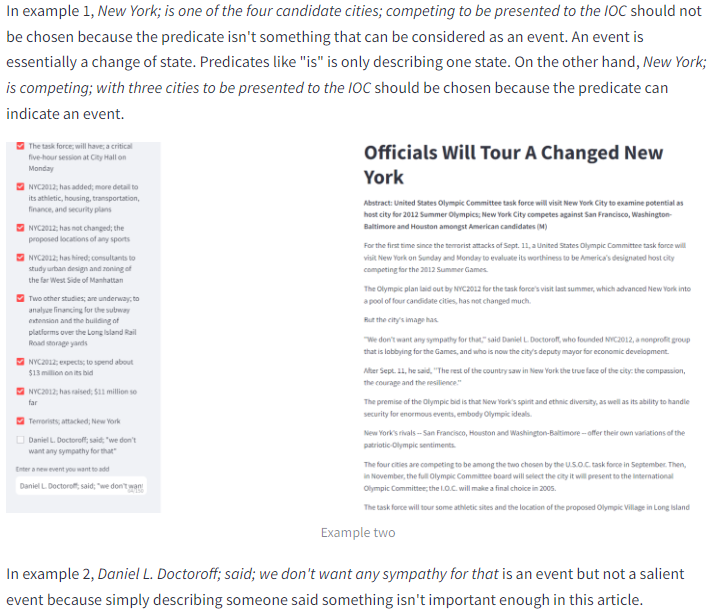}
\end{figure*}

\begin{figure*}
    \centering
    \includegraphics[width=0.6\linewidth]{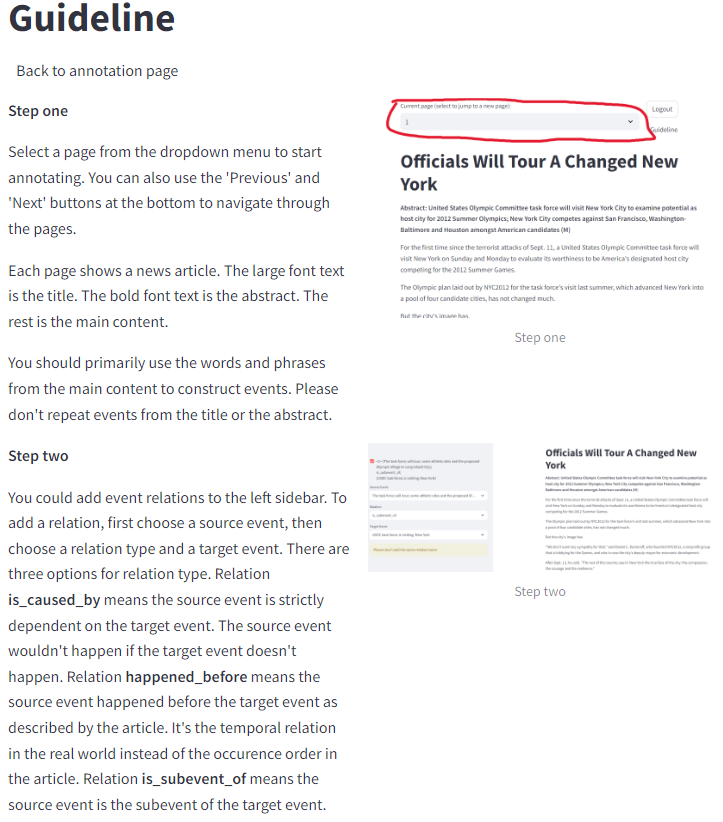}
\end{figure*}

\begin{figure*}
    \centering
    \includegraphics[width=0.6\linewidth]{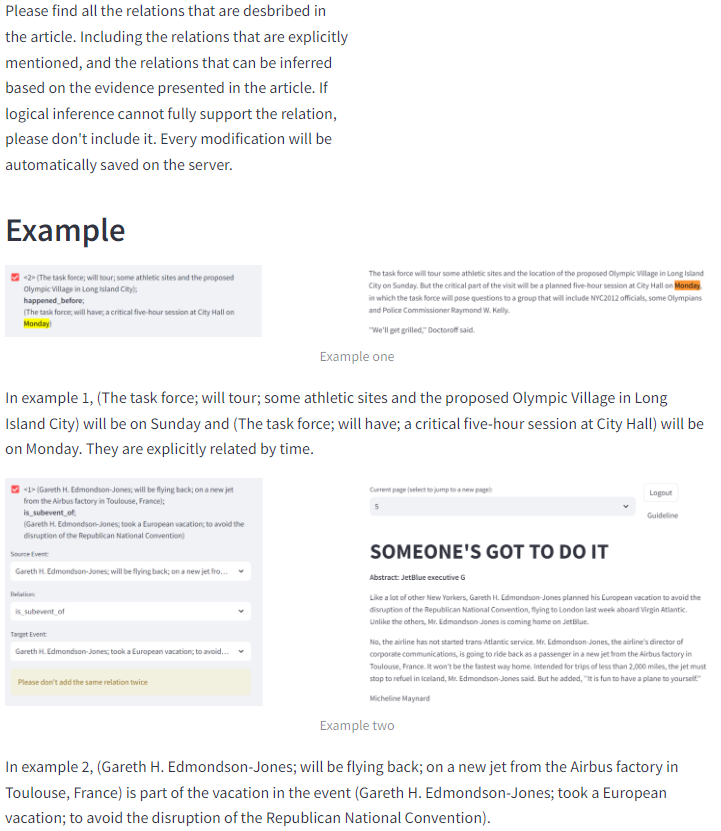}
    \caption{Annotation guidelines of relation identification shown to the annotators.}
    
\end{figure*}

\end{document}